# BiopSym : a simulator for enhanced learning of ultrasound-guided prostate biopsy

Stefano Sclaverano[1], Grégoire Chevreau[1,2], Lucile Vadcard[3], Pierre Mozer[2], Jocelyne Troccaz[1]

[1]*TIMC-IMAG Laboratory, UMR 5525 CNRS UJF, Grenoble, France*
[2]*Urology department, La Pitié Salpêtrière Hospital, Paris, France*
[3]*LSE, UPMF, Grenoble, France*

**Abstract**: This paper describes a simulator of ultrasound-guided prostate biopsies for cancer diagnosis. When performing biopsy series, the clinician has to move the ultrasound probe and to mentally integrate the real-time bi-dimensional images into a three-dimensional (3D) representation of the anatomical environment. Such a 3D representation is necessary to sample regularly the prostate in order to maximize the probability of detecting a cancer if any. To make the training of young physicians easier and faster we developed a simulator that combines images computed from three-dimensional ultrasound recorded data to haptic feedback. The paper presents the first version of this simulator.

## 1. Problem

Prostate cancer is the first cancer of men in many countries. When a cancer is suspected biopsies are collected in the gland, most often based on ultrasound (US) guidance, for further anatomo-pathologic analysis. The US probe is inserted into the rectum of the patient and a mechanical guide is attached to the probe for needle insertion (cf. fig.1). Based on prostate exploration and image understanding the clinician repeatedly positions the probe and collect a sample in order to execute a predefined sampling scheme. The clinician has to face several difficulties: (1) the cancer, if any, is generally not visible in the US images; (2) the biopsy schemes are 3D whilst US images are 2D; (3) the prostate may be moved and deformed by the US probe. Such difficulties limit the gesture accuracy and degrade the sensitivity of biopsies. We have developed a simulator for US-guided prostate biopsies (1) to enable relevant learning, (2) to make quantitative evaluation of operators possible and (3) to allow the quantitative comparisons of protocols.

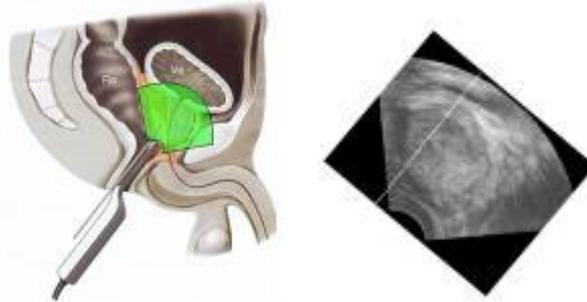

**Figure 1.** Prostate anatomy and US image acquisition for biopsy (left) – US image with needle trajectory (right).

In addition to the previous clinical difficulties, the training phase of this intervention also poses some problems. A large part of the expertise concerning biopsies relies on the coordination between vision (US 2D images), haptic perception (holding the probe) and cognition (mental representation of the gland). This coordination is necessary to link visual and haptic feedback to anatomical knowledge, and take decision concerning samples location. Simple observation of clinicians at work is not sufficient to provide residents with all these aspects of knowledge and to train their links in situation. In particular, the cognitive processes involved for the treatment visual information and for the control of actions (coordination of 2D and 3D) are far from being easy to elaborate [1], [2].

Simulation is a possible answer to this training difficulty. Simulators may involve role plays, virtual systems, or mixed systems including a virtual and a robotic part [3]. In any case, we argue that simulation alone is not sufficient to manage relevant training. Focusing on training implies to define some pedagogical features of the training system. Even in the case of *standardized patients*[1], these latter are aware of the kind of reaction and feedback they have to give residents during and at the end of the training session [4]. In the following sections we give details on the way we address these issues in the case of Biopsym.

**2.** Tools and methods

The simulator includes a haptic device, the Omni Phantom from Sensable Devices Inc., which allows the operator to move the virtual US probe with respect to the anatomy of the virtual patient (see fig.2). The system generates US images corresponding to the position of the probe. A haptic feedback renders both the constraint related to the motion of the probe in the rectum and the biopsy gun percussion when a sample is collected.

---

[1] Standardized patients are healthy volunteers who participate in a simulated examination. Of course, such type of simulation cannot be reasonably used for gestures such as biopsies.

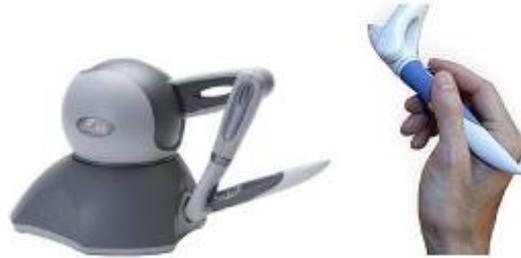

**Figure 2**: Using the Omni Phantom for the manipulation of the virtual US probe

*2.1. Methods*

Based on the motions of the virtual probe transmitted by the operator using the haptic device, the system generates corresponding US images. The haptic feedback renders both the constraint related to the motion of the probe in the rectum and the biopsy gun percussion when a sample is collected.

Two approaches can be used for US image simulation. One consists in generating US images [5] based on the modelling of US wave propagation and interaction with the human tissues. Such an approach generally requires complex models and is time consuming or images may have a limited realism when the model is simplified for computational efficiency. Another approach consists in using databases of recorded US exams and producing new images by exploiting this database. Interpolation may be necessary [6] when the image position does not correspond to recorded data. We have selected this second type of approach based on a 3D database. In this case, the generation of an image simply corresponds to slicing the volume in the direction of interest.

One major issue related to US images concerns the tissue deformations that may result from the image acquisition itself (probe pressure) or from external actions (needle insertions for instance). This issue not yet handled with this version of the simulator will be discussed in section 4. However, even without this function, the simulator may already help the trainee to navigate in the prostate volume, to visualize the prostate appearance and to correlate his/her probe 3D motions to corresponding 2D slices. We design targeted exercises for this training objective.

Biopsym uses open source libraries (see fig.3) and is multi-platform.

*2.2. Database*

3D US volumes have been acquired during real biopsy sessions at La Pitié Salpétrière Hospital for more than two years using a GE Voluson 730 system with a 3D endorectal US probe. By mid-june 2008, 87 patients were included. The database provides information about the prostate size, the patient age, the PSA level, etc. The SQL encoded database also includes information about operators using the simulator and records simulations of the operator in terms of patients concerned and positions of the simulated biopsies. Statistics can be computed from this database.

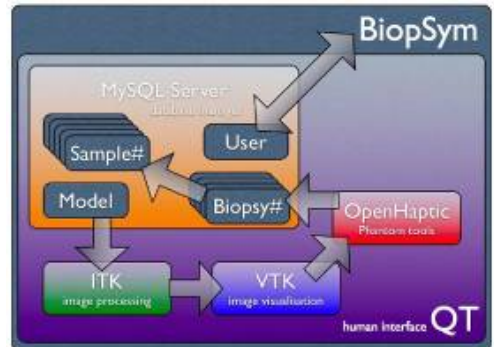
**Figure 3:** Software environment

### 3. Results

The simulator allows the trainee visualizing 3D US data to explore patients' anatomy using a multi-view rendering (axial, sagittal, coronal, probe-oriented 2D US images and 3D representation of the probe and needle with respect to the prostate volume) – see fig. 4.1. It also allows to simply visualize the 2D view corresponding to a real biopsy session and to exercise biopsies on the virtual patient (see fig. 4.2). This simulation includes the haptic feedback which makes the simulation more realistic. This haptic feedback can be switched off to make possible training on a PC with standard keyboard and mouse interfaces. Three clinicians have tested the simulator and they evaluate very positively this first version of the simulator from the viewpoint of image realism and haptic feeling.

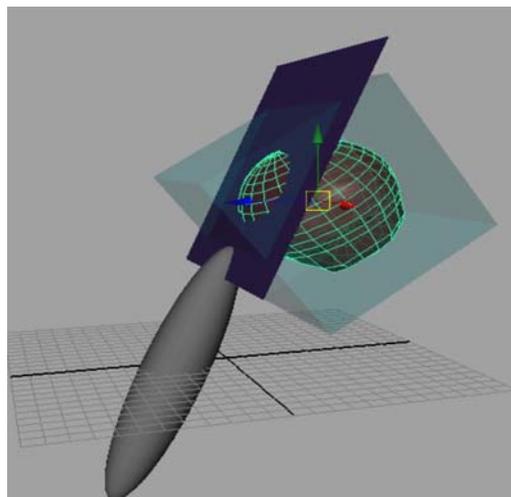
**Figure 4.1:** 3D visualization of the simulation environment (the probe and the plane to be visualized w.r.t. the volume of data and prostate).

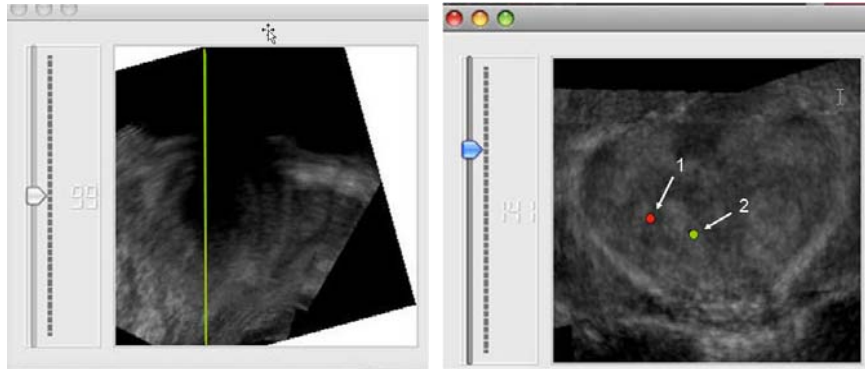

**Figure 4.2**: Multiple views of a biopsy trajectory computed from the 3D volume (left: sagittal – right: coronal – the needle trajectory is visualized in green in both views (line on left view and dot n°2 on right view); a previously recorded biopsy is in red in the right view (dot n°1)

Statistics computed from trainees' actions will be used to identify particular difficulties: for instance about sampling the apex of the prostate or its boundaries, about the length of biopsies (i.e. the actual depth of the needle in the gland), about the regular spreading of samples, etc. Those difficulties are linked to targeted sets of training exercises. Biopsym will thus offer personalised training paths, according to trainees' performances on the simulator.

## 4. Current and future work

On the scientific side, we plan to increase the image realism by adding real-time deformations of the images due to the probe. The generative approach can deal quite naturally with those questions provided that a model of deformation is available to deform the organs before image generation. Interpolation approaches may integrate deformed volumes to produce deformed images [6] or may apply a deformation model (mass-spring [7] or Finite-Element [8]) to images produced from a deformation-less recorded volume. We are in the process of comparing the advantages and drawbacks of the different approaches in the interpolation framework.

On the clinical side, further work deals with the extensive use of the simulator by clinicians with various levels of expertise. On the pedagogical side, the specific exercises will be included to help the learner in his/her skill acquisition: e.g. try to locate a given region in the 3D US volume or attempt to reach a specific target in the gland. Hints for visualisation will be available according to the trainee level (see fig. 4.1 for example). We will use data on learning curves (those of clinicians on real patients and those of trainees on Biopsym) to evaluate the impact of the simulator on biopsies performances.